%% file: main.tex
\documentclass[10pt,twocolumn,letterpaper]{article}

\usepackage{cvpr}              % To produce the CAMERA-READY version
\usepackage[accsupp]{axessibility}  % Improves PDF readability for those with disabilities.

\definecolor{cvprblue}{rgb}{0.21,0.49,0.74}
\usepackage[pagebackref,breaklinks,colorlinks,allcolors=cvprblue]{hyperref}
\usepackage{algorithm}
\usepackage{algorithmic}
\usepackage{amsmath}
\usepackage{xcolor}

\definecolor{grassgreen}{RGB}{34,139,34}

\def\paperID{4520} % *** Enter the Paper ID here
\def\confName{CVPR}
\def\confYear{2025}

\title{Modeling Multiple Normal Action Representations for Error Detection in Procedural Tasks}

\author{
    Wei-Jin Huang$^{1,*}$ \hspace{3mm} Yuan-Ming Li$^{1,*}$ \hspace{3mm}  Zhi-Wei Xia$^{1}$ \hspace{3mm}  Yu-Ming Tang$^{1}$ \hspace{3mm}  Kun-Yu Lin$^{1}$ \\
    Jian-Fang Hu$^{1}$ \hspace{3mm} Wei-Shi Zheng$^{1,2,3,4,\dagger}$ \\
    $^1$School of Computer Science and Engineering, Sun Yat-sen University, China \\
    $^2$Peng Cheng Laboratory, China\\
    $^3$Key Laboratory of Machine Intelligence and Advanced Computing, Ministry of Education, China; \\
    $^4$Guangdong Province Key Laboratory of Information Security Technology, China\\
    {\tt\small \{huangwj235, liym266\}@mail2.sysu.edu.cn; wszheng@ieee.org }
}
\usepackage{pifont} % 导入 pifont 包
\usepackage{multirow} % 如果需要使用 \multirow
\usepackage{colortbl}
\usepackage{xcolor}
\newcommand{\cmark}{\ding{51}} % 定义打勾符号
\newcommand{\xmark}{\ding{55}} % 定义打叉符号

\begin{document}
\maketitle

% \author{\authorBlock}
\maketitle
{\let\thefootnote\relax\footnotetext{
\scriptsize 
*: Equal contributions. {$\dagger$}: Corresponding author. 
}}

\begin{abstract}
Error detection in procedural activities is essential for consistent and correct outcomes in AR-assisted and robotic systems. Existing methods often focus on temporal ordering errors or rely on static prototypes to represent normal actions. However, these approaches typically overlook the common scenario where multiple, distinct actions are valid following a given sequence of executed actions. This leads to two issues: (1) the model cannot effectively detect errors using static prototypes when the inference environment or action execution distribution differs from training; and (2) the model may also use the wrong prototypes to detect errors if the ongoing action label is not the same as the predicted one. To address this problem, we propose an Adaptive Multiple Normal Action Representation (AMNAR) framework. AMNAR predicts all valid next actions and reconstructs their corresponding normal action representations, which are compared against the ongoing action to detect errors. Extensive experiments demonstrate that AMNAR achieves state-of-the-art performance, highlighting the effectiveness of AMNAR and the importance of modeling multiple valid next actions in error detection. The code is available at \url{https://github.com/iSEE-Laboratory/AMNAR}.
\end{abstract}

\section{Introduction}
\label{sec:intro}
Understanding procedural activities is an important field in video action understanding \cite{li2024continual,li2024techcoach,lin2024human,zhou2024mitigating,lin2024rethinking,zhou2024actionhub,lin2023diversifying}, as it reflects how an AI model recognizes actions \cite{carreira2017quo, wang2018non, weng2023open,zeng2020hybrid,xu2022likert,zeng2024multimodal}, separates steps \cite{cheng2014temporal, richard2016temporal, lea2017temporal, farha2019ms, li2020ms, yi2021asformer, zhang2022actionformer}, and plans movements \cite{guo2024uncertainty, gong2022future, wang2023event,mittal2024can}, similar to how humans conduct daily tasks (e.g., cooking, assembling toys, using tools, etc.). Since procedural tasks require consistent outcomes without errors, being able to detect natural mistakes is a crucial ability for next-generation AR assisted and robotic systems \cite{wang2023holoassist,duan2024aha,grauman2024ego,wang2025task,zheng2024selective,zheng2025diffuvolume}.

\begin{figure}[t]
    \centering
    \includegraphics[width=1.0\linewidth]{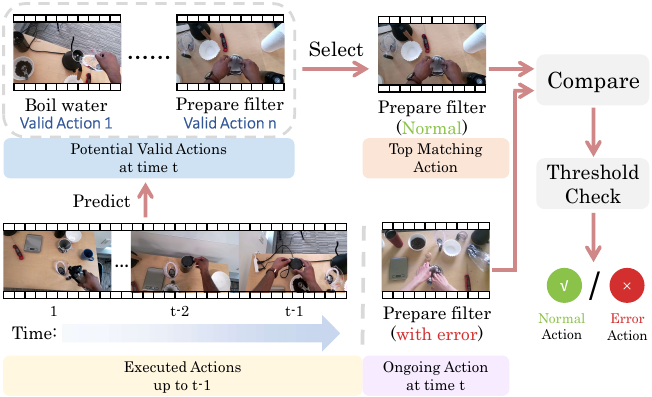}
    \vspace{-0.6cm}
    \caption{\textbf{Illustration of error detection using multiple valid next actions at time \( t \).} After ``Grinding Coffee Bean'' at time \( t-1 \), valid next actions include ``Boil Water'' and ``Prepare Filter.'' The best matching action is selected and compared with the ongoing action. If their distance exceeds the threshold, the action is marked as an error; otherwise, it is marked as normal.}
    \label{fig:valid_actions}
    \vspace{-15pt}
\end{figure}

Achieving reliable error detection requires a foundational understanding of what a normal action should be, allowing comparison to the ongoing action. So, for error detection using AI, a key challenge is constructing a ``normal action representation" of the current action. Approximating a ``normal action representation," some current methods \cite{seminara2024differentiable, ding2023every, flaborea2024prego} model transition relationships from correctly executed action sequences, thereby predicting the correct current action label based on prior actions. However, while these approaches can detect label-level errors by comparing predicted and actual action labels, they often fail to capture the full representation of a normal action, leading them to overlook cases where the correct action label is executed but deviates from the expected normal behavior. For instance, pouring water into a filter but spilling some outside shares the same label as the correct action, yet introduces an error (spillage) that current methods often miss.

Recently, a contrastive learning method\cite{lee2024error} tackles this limitation by leveraging prototype-based representations. This method learns a series of prototypes during training to represent the normal execution of each action class. 
During inference, it detects errors by comparing the current action’s representation to the closest prototype in the same class. 
While effective in some scenarios, these prototypes are static after the training stage and cannot adaptively change with action executions, which struggles to detect errors effectively when the action distribution varies from that of the training samples (e.g., diverse execution styles, tools with varied appearances, or distinct environments).

To detect errors with varying action distributions, we argue that an effective error detection model should be capable of dynamically generating a normal action representation for the current action, conditioned on previously executed actions. Such adaptability allows the model to account for variations in action execution that static prototypes cannot handle. From this perspective, a straightforward implementation might attempt to predict a normal action representation of the current action from the past action sequence. However, this naive approach introduces a new challenge: after a sequence of executed actions, multiple valid next actions may logically follow, depending on user preferences or contextual factors, rather than adhering to a strictly defined sequence. For instance, in the process of making coffee, after grinding the coffee beans, the next step might involve boiling water, preparing the filter, or selecting a cup—each of which is a valid action depending on the task context, as illustrated in \cref{fig:valid_actions}. This range of valid next actions following a given sequence of actions makes it challenging for a single normal action representation to encompass all possible correct action representations.

\begin{figure*}[t]
    \centering
    \includegraphics[width=0.95\linewidth]{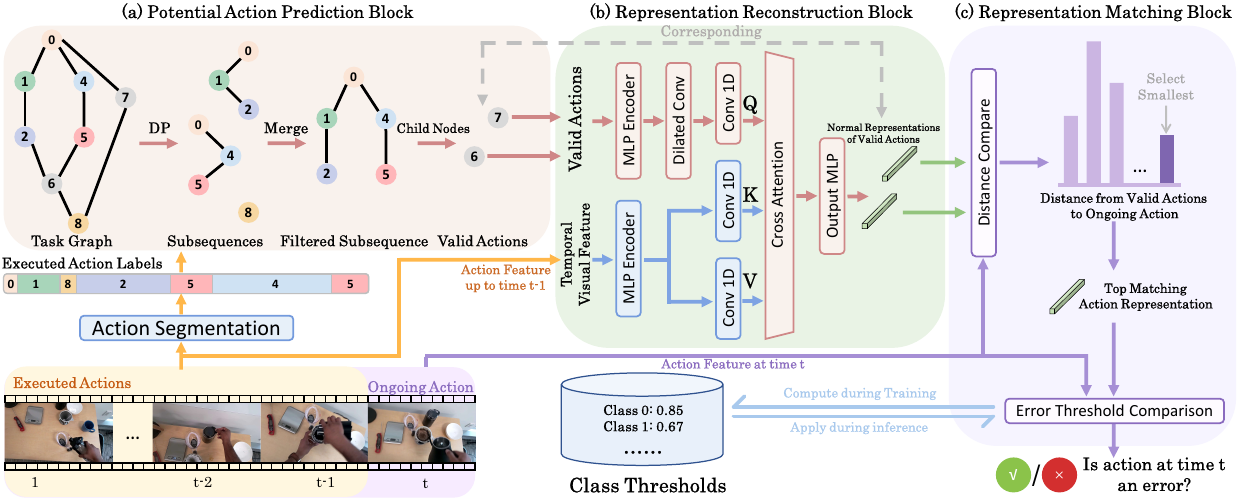} 
    \vspace{-0.3cm}
    \caption{\textbf{Overview of the Adaptive Multiple Normal Action Representation (AMNAR) framework.} The process begins with an Action Segmentation module identifying executed actions from video input. \textbf{(a) Potential Action Prediction Block} predicts valid next actions using a task graph from executed action labels. \textbf{(b) Representation Reconstruction Block} generates normal action representations for these valid actions, leveraging temporal visual features. \textbf{(c) Representations Matching Block} compares the ongoing action’s feature at time t with the generated representations to detect errors, indicated by a checkmark (\cmark) for normal actions or a cross (\xmark) for errors.}
    \label{fig:method_framework}
    \vspace{-0.6cm}
\end{figure*}

To overcome these challenges, we propose a novel Adaptive Multiple Normal Action Representation (AMNAR) framework for error detection that predicts and models multiple valid next actions, dynamically creating multiple adaptable normal action representations that accurately capture the diversity of procedural task execution. Specifically, our approach first employs an action segmentation model to provide an initial executed action sequence, which serves as input for the Potential Action Prediction Block (PAPB). Subsequently, PAPB predicts all valid actions based on executed actions leveraging the task graph and dynamic programming. Next, our Representations Reconstruction Block (RRB) reconstructs multiple normal action representations for each valid action. As shown in \cref{fig:method_framework}, these modules enable AMNAR to represent all valid actions adaptively in error detection. Finally, the Representation Matching Block (RMB) selects the most likely normal action representation of current execution step and assesses its conformity with the ongoing action. By comparing this conformity to a predefined threshold, we determine whether the current action has an error.

% To evaluate the effectiveness of our proposed error detection framework, we conduct experiments on three datasets: EgoPER \cite{lee2024error}, HoloAssist \cite{wang2023holoassist} and CaptainCook4D\cite{peddi2023captaincook4d}. Our experimental results demonstrate that our method achieves an average improvement of \textbf{7.4\%} in Error Detection Accuracy (EDA) and \textbf{6.5\%} in Area Under the Curve (AUC) on the EgoPER dataset. For the HoloAssist dataset, our approach yields an average increase of \textbf{3.7\%} in EDA and \textbf{1.3\%} in AUC. Furthermore, ablation studies on our framework validate the contributions of each component, revealing several insights.

We evaluate our error detection framework on three datasets: EgoPER \cite{lee2024error}, HoloAssist \cite{wang2023holoassist}, and CaptainCook4D \cite{peddi2023captaincook4d}. Our method improves Error Detection Accuracy (EDA) by \textbf{7.4\%} and Area Under the Curve (AUC) by \textbf{6.5\%} on EgoPER, \textbf{3.7\%} and \textbf{1.3\%} on HoloAssist, and \textbf{2.5\%} and \textbf{5.3\%} on CaptainCook4D, proving its robustness across tasks. Furthermore, ablation studies on our framework validate the contributions of each component, revealing several insights.

The main contributions of our work are as follows:
\begin{enumerate}
    \item We introduce a novel approach that dynamically generates multiple normal action representations for the current action, conditioned on previously executed actions, addressing the challenge of adaptively representing all normal action representations of the current action.
    \item We develop a new framework utilizing task graphs and dynamic programming to predict multiple valid next actions. This enables precise and context-aware error detection by comparing the current action with adaptively generated multiple normal action representations.
    \item Comprehensive experiments on the EgoPER, HoloAssist and CaptainCook4D datasets validate the effectiveness of each component in our framework, achieving state-of-the-art performance and demonstrating robustness and flexibility in handling diverse procedural task errors.
\end{enumerate}

%  ========== RELATED WORKS
\section{Related Work}
\label{sec:related}
% In this section, we will introduce the relevant work in the field of Error Detection, as well as the related field to our work: Video Anomaly Detection.
% \subsection{Error Detection}
\noindent\textbf{Error Detection }\cite{ghoddoosian2023weakly,peddiput,schoonbeek2024industreal,wang2023holoassist,jang2019epic,li2024egoexo,grauman2024ego,peddi2023captaincook4d,sener2022assembly101,flaborea2024prego,ding2023every,lee2024error,seminara2024differentiable,haneji2024egooops,mazzamuto2024eyes} is a task to detect when errors occur in procedural activities. Starting from EPIC-Tent\cite{sener2022assembly101}, many researchers have begun to focus on the Error Detection task. Ding et al.\cite{ding2023every} attempt to identify both order-related errors in toy assembly and part-relative placement errors by solving the Error Detection problem using the constructed Graphs method. Subsequently, several works \cite{flaborea2024prego,seminara2024differentiable} focus on detecting order-related errors. Specifically, Flaborea et al.\cite{flaborea2024prego} focus on utilizing large language models to identify order-related errors, while Seminara et al.\cite{seminara2024differentiable} propose using differentiable task graph matrices for the same purpose.
Recently, EgoPED \cite{lee2024error} expands this field by introducing errors beyond action order (\eg, omission, addition or modification of steps) and proposes a new contrastive step prototype learning framework. Our work shares the same problem formulation with EgoPED \cite{lee2024error}. Differently, we address the challenge of multiple valid next actions that can follow any given step in procedural tasks. Our approach adaptively models all normal action representations of these valid actions, enabling robust error detection through adaptively generated normal action representations for each possible action.

\vspace{0.1cm}

\noindent\textbf{Video Anomaly Detection (VAD)} focuses on spotting unusual events (\eg, accidents or suspicious behaviors) in surveillance footage. VAD identifies deviations from normal activities in videos, which could indicate dangerous or unexpected situations such as falls or unauthorized entries into restricted areas. One of the main branches in this field is the reconstruction-based VAD \cite{liu2021hybrid,gong2019memorizing,yang2023video,ristea2022self,ristea2024self}. In this branch, the models are trained to reconstruct normal frames or accurate sequences, and detect the anomaly by measuring the reconstruction error between the reconstructed and original frames (or sequences).
Unlike VAD, which primarily detects deviations based on low-level visual or statistical anomalies, our approach assesses whether ongoing actions align with the predicted normal action representations of all valid next actions. Additionally, while VAD typically reconstructs a single normal scene, our method addresses the challenge of error detection in complex procedural tasks by simultaneously modeling multiple normal action representations for multiple valid next actions.

\section{Method}
\vspace{-0.1cm}
\label{sec:method}
To adaptively reconstruct all normal action representations of valid actions, which are compared with ongoing actual action to detect errors, we propose an Adaptive Multiple Normal Action Representation (AMNAR) framework that explicitly models diverse normal action representations of all valid next actions based on executed action sequence.

We will first introduce problem formulation in \cref{sec:problem_formulation}. 
After that, we provide an overview of the proposed method AMNAR in \cref{sec:method_overview}, and introduce the detailed designs in \cref{sec:executed_action_sequences_and_action_features,sec:potential_action_prediction,sec:representation_reconstruction,sec:normal_action_representation_alignment_conformity_assessment,sec:training_strategy_and_objective_function}.

\subsection{Problem Formulation}
\vspace{-0.1cm}
\label{sec:problem_formulation}
Following previous work\cite{lee2024error}, we train our model with normal videos and their corresponding action labels for each procedural task execution step. Each video, denoted as \( V = \{f_i\}_{i=1}^N \), is paired with frame-wise action labels \( Y = \{y_i\}_{i=1}^N \), where $N$ indicates the number of frames, and each \( y_i \) maps to one of \( S \) predefined action classes or to a background class, expressed as \( y_i \in \{1, 2, \ldots, S, S+1\} \). During inference, the objective is to detect error actions, denoted as \( E = \left\{e_j\right\}_{j=1}^M \), where \( M \) is the number of errors.

\subsection{Method Overview}
\vspace{-0.1cm}
\label{sec:method_overview}
Given that multiple valid next actions may follow a given executed action sequence, we need to create accurate representations for each possible valid action to detect small deviations in how actions are executed, even if the overall action type is correct. Combining all valid actions with video context, our method reconstructs all normal action representations for each possible current action. The top matching representation to the actual action is then selected and used to detect errors. To do this, we propose an Adaptive Multiple Normal Action Representation (AMNAR) framework, as illustrated in \cref{fig:method_framework}.

AMNAR initially extracts visual features from videos using a visual feature extractor. The features are then fed into an Action Segmentation Model, which outputs labeled action segments with start and end frames. Subsequently, the \textbf{Potential Action Prediction Block} predicts all valid next actions based on the task graph and executed actions. The \textbf{Representation Reconstruction Block} then generates representations for these predicted actions. Lastly, the \textbf{Representation Matching Block} assesses any deviations between these representations and the ongoing action features to identify possible errors in the ongoing action.

% \subsection{Executed Action Sequences and Action Features}
\subsection{Action Sequences and Features Execution}
\vspace{-0.1cm}
\label{sec:executed_action_sequences_and_action_features}
We represent procedural task actions using a pre-trained feature extractor (e.g., I3D \cite{carreira2017quo}) to obtain initial visual features, which are processed by the Action Segmentation Model (ASM). The ASM identifies action segments \( A = \{a_k\}_{k=1}^H = \{(y_k, st_k, ed_k)\}_{k=1}^H \), where \( H \) is the total number of actions, \( y_k \) is the label for the \( k \)-th segment, and \( st_k \) and \( ed_k \) denote the start and end frames. The ASM also generates a refined feature set \( F = \{f_i\}_{i=1}^N \), with each \( f_i \) as a frame-level feature vector.

The executed action sequence up to time \( t \), denoted \( s_t = \{y_k\}_{k=1}^{t-1} \), captures prior actions. To account for varying segment lengths, we compute an action feature \( f_t^{\text{action}} \) by averaging frame-level features within each segment:
\vspace{-0.2cm}
\begin{equation}
   f_t^{\text{action}} = \text{average}(\{f_i\}_{i=st_t}^{ed_t}),
   \vspace{-0.2cm}
\end{equation}
where the average is taken over frames from \( st_t \) to \( ed_t \).

% \vspace{-0.1cm}
\subsection{Potential Action Prediction Block}
\vspace{-0.1cm}
\label{sec:potential_action_prediction}

In procedural tasks, multiple valid actions may follow a given execution sequence. To address this, our Potential Action Prediction Block (PAPB) identifies all valid next steps using a predefined task graph \( G \) that encodes task-specific action sequences. PAPB maps the current action sequence \( s_t = \{y_1, y_2, \dots, y_{t-1}\} \), where each \( y_i \) represents an executed action label, onto \( G \) to determine the set of logically valid next actions.

To handle potential inaccuracies of action segmentation label, such as mislabeled or omitted actions, PAPB employs dynamic programming (DP) to compute \( s_{t}^* \), the filtered subsequence of \( s_t \) that aligns with \( G \), as illustrated in part (a) of \cref{fig:method_framework} and \cref{fig:task_graph}. For example, given an executed sequence \( s_t = [0, 1, 8, 2, 5, 4, 5] \). Using DP, PAPB identifies the longest common subsequences (lcs) that form non-branching paths in \( G \). It maintains two arrays: \( \text{dp}[i] \), tracking the length of the longest non-branching subsequence ending at index \( i \), and \( \text{subseq}[i] \), storing the corresponding subsequence. Updates occur as follows:

\vspace{-0.2cm}
\begin{equation}
\text{dp}[i] = \max(\text{dp}[i], \text{dp}[j] + 1),
\vspace{-0.4cm}
\end{equation}
\begin{equation}
\resizebox{0.9\linewidth}{!}{$
\text{subseq}[i] =
\begin{cases} 
    \text{subseq}[j] \cup \{y_i\}, & \text{if } \text{dp}[j] + 1 > \text{dp}[i], \\
    \text{subseq}[i] \cup (\text{subseq}[j] \cup \{y_i\}), & \text{if } \text{dp}[j] + 1 = \text{dp}[i],
\end{cases}
$}
\vspace{-0.2cm}
\end{equation}
where a non-branching subsequence is a continuous path in \( G \) without splits (e.g., [0, 1, 2] in the task graph).

Next, PAPB merges these subsequences into a unified subgraph, as shown in part (a) of \cref{fig:method_framework} and \cref{fig:task_graph}. Since node 0 is shared between \( \text{lcs}_1 \) and \( \text{lcs}_2 \), they are merged into \( s_{t}^* = [0, 1, 2, 4, 5] \), capturing all relevant executed actions. Finally, PAPB identifies the valid next actions \( C_t \) by extracting the child nodes of \( s_{t}^* \) in \( G \):
\vspace{-0.2cm}
\begin{equation}
C_t = (\bigcup_{a \in s_{t}^*} A[a]) \setminus s_{t}^*,
\vspace{-0.2cm}
\end{equation}
where \( \bigcup_{a \in s_{t}^*} A[a] \) aggregates all successors of nodes in \( s_{t}^* \) from the adjacency list A of task graph G, and \( \setminus s_{t}^* \) excludes already executed actions. For instance, in \cref{fig:task_graph}, the child nodes of \( s_{t}^* \) include nodes 6 and 7, \( C_t = [6, 7] \). This process ensures \( C_t \) robustly represents all potential valid next actions, enhancing error detection in complex procedural workflows. More details can be found in supplementary.

\begin{figure}[t]
    \centering
    \includegraphics[width=0.95\linewidth]{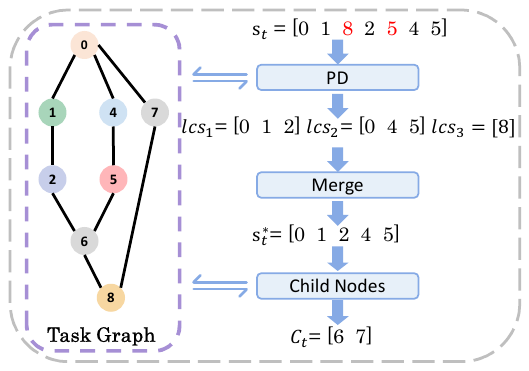} 
    \vspace{-0.3cm}
    \caption{\textbf{Overview of the Potential Action Prediction Block.} Using Dynamic Programming (DP), this module identifies all longest common subsequences (lcs) from the executed action sequence \( s_t \) via the task graph \( G \). These lcs are interconnected into a unified subgraph, forming the filtered sequence \( s_t^* \). Reachable child nodes from \( G \) are then extracted as valid next actions \( C_t \).}
    \label{fig:task_graph}
    \vspace{-0.3cm}
\end{figure}

\subsection{Representation Reconstruction Block}
\vspace{-0.1cm}
\label{sec:representation_reconstruction}

We propose the Representation Reconstruction Block (RRB) to generate normal action representations for valid next actions, as shown in part (b) of \cref{fig:method_framework}. Starting with frame-wise features \( F_{1:ed_{t-1}} = \{ f_i \}_{i=1}^{ed_{t-1}} \) up to time \( t \), we apply dilated convolution to capture long-range dependencies, yielding \( F_{1:ed_{t-1}}^{\text{conv}} \).

Then, we use local cross-attention to align the valid actions with the temporal context. Here, \( C_t \) serves as the query, while \( F_{1:ed_{t-1}}^{\text{conv}} \) provides the keys and values. This mechanism generates a contextual feature for each valid action \( y_{t,i} \in C_t \). To enhance robustness, we adopt a cluster-residual prediction approach. Specifically, for each valid action \( y_{t,i} \), we predict a residual \( r_{t,i} \) that refines the cluster center \( c_{t,i} \) of the corresponding action class. The normal action representation \( f_{t,i}^{\text{normal}} \) is then computed as:
\vspace{-0.1cm}
\begin{equation}
    r_{t,i} = \mathcal{E}_{\text{res}}(y_{t,i}, F_{1:ed_{t-1}}^{\text{conv}}),
\vspace{-0.4cm}
\end{equation}
\begin{equation}
    f_{t,i}^{\text{normal}} = c_{t,i} + r_{t,i},
\vspace{-0.1cm}
\end{equation}
where \( \mathcal{E}_{\text{res}} \) denotes the residual prediction operation via cross-attention, and \( c_{t,i} \) is the precomputed cluster center for action class \( y_{t,i} \), derived from normal training samples. The resulting set of representations, denoted \( f_t^{\text{normal}} = \{ f_{t,i}^{\text{normal}} \}_{i=1}^{|C_t|} \), encapsulates all valid next actions:
\vspace{-0.2cm}
\begin{equation}
    f_t^{\text{normal}} = \mathcal{E}(C_t, F_{1:ed_{t-1}}),
\vspace{-0.2cm}
\end{equation}
where \( \mathcal{E} \) integrates all operations within this block. This approach leverages the cluster center as a stable baseline while adapting to context-specific variations through the residual.

\begin{table*}[h]
\centering
\caption{\textbf{Comparison with existing methods on the EgoPER dataset for each task and the average over all tasks.}}
\vspace{-0.2cm}
\resizebox{0.95\textwidth}{!}{ % Resize table to fit page width
\begin{tabular}{l| cc cc cc cc cc |cc}
\toprule
\multirow{2}{*}{Methods} & \multicolumn{2}{c}{Quesadilla} & \multicolumn{2}{c}{Oatmeal} & \multicolumn{2}{c}{Pinwheel} & \multicolumn{2}{c}{Coffee} & \multicolumn{2}{c|}{Tea} & \multicolumn{2}{c}{All} \\
 & EDA & AUC & EDA & AUC & EDA & AUC & EDA & AUC & EDA & AUC & EDA & AUC \\
\midrule
Random & 19.9 & 50.0 & 11.8 & 50.0 & 15.7 & 50.0 & 8.20 & 50.0 & 17.0 & 50.0 & 14.5 & 50.0 \\
HF$^2$-VAD \cite{liu2021hybrid} & 34.5 & 62.6 & 25.4 & 62.3 & 29.1 & 52.7 & 10.0 & 59.6 & 36.6 & 62.1 & 27.1 & 59.9 \\
HF$^2$-VAD + SSPCAB \cite{ristea2022self} & 30.4 & 60.9 & 25.3 & 61.9 & 33.9 & 51.7 & 10.0 & 60.1 & 35.4 & 63.2 & 27.0 & 59.6 \\
S3R \cite{wu2022self} & 52.6 & 51.8 & 47.8 & 61.6 & 50.5 & 52.4 & 16.3 & 51.0 & 47.8 & 57.9 & 43.0 & 54.9 \\
EgoPED\cite{lee2024error} & \textbf{62.7} & 65.6 & 51.4 & 65.1 & 59.6 & 55.6 & 55.3 & 58.3 & 56.0 & \textbf{66.0} & 57.0 & 62.0 \\
\midrule
AMNAR (Ours) & 61.4 & \textbf{71.9} & \textbf{65.0} & \textbf{75.4} & \textbf{65.0} & \textbf{65.4} & \textbf{73.5} & \textbf{67.8} & \textbf{57.0} & 61.9 & \textbf{64.4} & \textbf{68.5} \\
\bottomrule
\end{tabular}
}
\label{tab:error_detection_egoper}
\vspace{-13pt}
\end{table*}

\subsection{Normal Action Representation Alignment and Conformity Assessment}
\vspace{-0.1cm}
\label{sec:normal_action_representation_alignment_conformity_assessment}

To assess action conformity and detect potential errors, we propose the Representation Matching Block (RMB), as illustrated in part (c) of \cref{fig:method_framework}. This component measures the deviation between normal action representations \( f_t^{\text{normal}} \) and ongoing action features \( f_t^{\text{action}} \). Specifically, RMB evaluates the alignment between the ongoing action feature \( f_t^{\text{action}} \) and each potential normal action representation feature \( f_{t,i}^{\text{normal}} \) for valid next actions. We calculate the Euclidean distance \( d_{t,i} \) as follows:
\vspace{-0.2cm}
\begin{equation}
d_{t,i} = \| f_t^{\text{action}} - f_{t,i}^{\text{normal}} \|_2.
\vspace{-0.2cm}
\end{equation}
To identify the best alignment, we select the smallest of these distances (\ie, \( d_t^{\text{min}} = \min_{i} d_{t,i} \)) as the top matching of ongoing action to any normal action representation.

\vspace{-0.2cm}
\paragraph{Error Detection Criterion.}
\label{error_detection_criterion}
To determine if an ongoing action deviates from all potential normal action representations, we apply a threshold \( \theta(y_t) \) specific to each action class \( y_t \), based on the distribution of alignment distances in normal samples. An action is flagged as an error if:
\vspace{-0.2cm}
\begin{equation}
\text{Error} =
\begin{cases}
    1, & \text{if } d_t^{\text{min}} > \theta(y_t), \\
    0, & \text{otherwise}.
\end{cases}
\vspace{-0.2cm}
\end{equation}
If \( d_t^{\text{min}} \) exceeds \( \theta(y_t) \), this indicates that the ongoing action at time \( t \) does not conform to any normal action representation, thereby identifying it as an error. If it remains within the threshold, the ongoing action is considered to conform to an expected, normal action.

\subsection{Training Strategy and Objective Function}
\vspace{-0.1cm}
\label{sec:training_strategy_and_objective_function}

In this section, we outline the model training process and our objective function.

During training, segments predicted by the Action Segmentation Model may include inaccuracies, such as incomplete or incorrect boundaries. To filter these samples, we calculate an overlap ratio \( R_{\text{overlap}} \) between each predicted segment \( S_{\text{pred}} \) and its closest ground truth segment \( S_{\text{GT}} \). The overlap ratio \( R_{\text{overlap}} \) is defined as:

\vspace{-0.2cm}
\begin{equation}
   R_{\text{overlap}} = \frac{| S_{\text{pred}} \cap S_{\text{GT}} |}{| S_{\text{pred}} |}.
\vspace{-0.2cm}
\end{equation}

We filter the segment based on whether \( R_{\text{overlap}} \) meets a predefined threshold \( \tau \):
\vspace{-0.2cm}
\begin{equation}
   S_{\text{filtered}} = 
   \begin{cases} 
      0, & \text{if } R_{\text{overlap}} < \tau \quad \text{(exclude)}, \\
      1, & \text{if } R_{\text{overlap}} \geq \tau \quad \text{(retain)}.
   \end{cases}
\vspace{-0.2cm}
\end{equation}

For retained segments, the assigned label may still need refinement. We assign each action segment a representative label \( y_t^* \) by selecting the most frequently occurring ground truth label within the frames of the segment. Let
\vspace{-0.2cm}
\begin{equation}
   L_t = \{ y_i^{\text{GT}} \mid i \in \text{segment} \},
\vspace{-0.2cm}
\end{equation}
where \( y_i^{\text{GT}} \) is the ground truth label for the \( i \)-th frame within the segment. Then, we define \( y_t^* \) as:
\vspace{-0.2cm}
\begin{equation}
   y_t^* = \arg\max_{y} \text{count}(y, L_t),
\vspace{-0.2cm}
\end{equation}
where \( \arg\max \) identifies the label \( y \) that appears most frequently in \( L_t \).

Since each training sample video corresponds to a single target action at time \( t \) in \( C_t \), we omit the Potential Action Prediction Block (PAPB) and set the query size of the Representation Reconstruction Block (RRB) to one during training. Using features from \cref{sec:executed_action_sequences_and_action_features}, we obtain frame-wise features \( F_{1:ed_{t-1}} = \{ f_i \}_{i=1}^{ed_{t-1}} \) up to time \( t \), as well as the action feature \( f_t^{\text{action}} \). Based on \( C_t \) and \( F_{1:ed_{t-1}} \), our method then predicts an normal action representation vector \( f_t^{\text{normal}} \) for the action at time \( t \):
\vspace{-0.2cm}
\begin{equation}
   f_t^{\text{normal}} = \mathcal{E}(C_t, F_{1:ed_{t-1}}),
\vspace{-0.2cm}
\end{equation}
where \( \mathcal{E} \) represents the operations within the Representation Reconstruction Block (\cref{sec:executed_action_sequences_and_action_features}).

Since \( C_t \) contains only one action during training, \( f_t^{\text{normal}} \) includes only one expected representation, \( f_{t,1}^{\text{normal}} \). Our optimization objective then minimizes the distance between this normal action representation and the actual action feature \( f_t^{\text{action}} \), as shown below:
\vspace{-0.2cm}
\begin{equation}
   \mathcal{L}_{\text{normal}} = \left\| f_{t,1}^{\text{normal}} - f_t^{\text{action}} \right\|^2.
\vspace{-0.2cm}
\end{equation}

\section{Experiments}
\vspace{-0.1cm}
\label{sec:experiments}

In this section, we first introduce our experimental setup in \cref{sec:experimental_setup}, followed by the evaluation metrics used to assess the performance of our method in \cref{sec:results_and_comparison}. Finally, we conduct ablation studies in \cref{sec:effectiveness_of_AMNAR_component} to demonstrate the effectiveness of our proposed AMNAR framework.

\subsection{Experimental Setup}
\vspace{-0.1cm}
\label{sec:experimental_setup}

\noindent\textbf{Datasets.}
We conduct experiments on the EgoPER \cite{lee2024error}, HoloAssist \cite{wang2023holoassist} and CaptainCook4D\cite{peddi2023captaincook4d} datasets. \textbf{The EgoPER dataset} is an egocentric video dataset with five cooking tasks (\ie, Quesadilla, Qatmeal, Pinwheel, Coffee and Tea). It includes 386 videos (28 hours) with both normal and erroneous executions, together with frame-level action and error annotations.
\textbf{The HoloAssist dataset} features 166 hours of video from 350 instructor-performer pairs completing real-world tasks (e.g., furniture assembly, device operation).
\textbf{The CaptainCook4D dataset} is an egocentric dataset of cooking activities, comprising 384 recordings of 24 recipes. It captures both normal and error executions, with step annotations and seven error types.

\begin{table}[t]
\centering
\caption{\textbf{Comparison on HoloAssist dataset.} Lacking action segmentation annotations, we train with fine-grained noun and verb labels. “Noun” and “Verb” denote ASM training with only noun or verb annotations, respectively; “All” averages both results.}
\vspace{-0.2cm}
\resizebox{\columnwidth}{!}{
\begin{tabular}{l| cc cc |cc}
\toprule
\multirow{2}{*}{Methods} & \multicolumn{2}{c}{Noun} & \multicolumn{2}{c|}{Verb} & \multicolumn{2}{c}{All} \\
 & EDA & AUC & EDA & AUC & EDA & AUC \\
\midrule
Random & 48.3 & 52.4 & 49.8 & 48.7 & 49.0 & 50.6 \\
EgoPED\cite{lee2024error} & 65.2 & 56.1 & 67.2 & 54.3 & 66.2 & 55.2 \\
\midrule
AMNAR (Ours) & \textbf{67.2} & \textbf{56.8} & \textbf{72.6} & \textbf{56.2} & \textbf{69.9} & \textbf{56.5} \\
\bottomrule
\end{tabular}
}
\label{tab:error_detection_results_in_holoassist}
\vspace{-10pt}
\end{table}

\noindent\textbf{Evaluation Metrics.}
Following prior work \cite{lee2024error}, we evaluate models with \textbf{Error Detection Accuracy (EDA)} and \textbf{Area Under the Curve (AUC)}. \textbf{EDA} measures the accuracy in identifying both erroneous and normal segments, reflecting the model’s overall accuracy in error detection at the segment level. \textbf{AUC} evaluates the ability of the model to distinguish between errors and non-errors by comparing true and false positive rates across varying thresholds.

\vspace{0.1cm}
\noindent\textbf{Implementation Details.}
Following EgoPED\cite{lee2024error}, we adopt I3D \cite{carreira2017quo} for video feature extraction and ActionFormer \cite{zhang2022actionformer} for action segmentation, with the segmentation model pretrained before joint training with other components. For EgoPER, we use provided task graphs, while for HoloAssist and CaptainCook4D, we construct task graphs from training sequences. In the Representation Reconstruction Block (\cref{sec:representation_reconstruction}), actions in \( C_t \) are represented by their class cluster centers. The error detection threshold \( \theta(y_t) \) (\cref{error_detection_criterion}) is set at the 0.85 quantile of the distance distribution from normal training instances, with an overlap threshold \( \tau = 0.6 \). More details are in the supplementary.

\begin{table}[t]
\centering
\caption{\textbf{Comparison on the CaptainCook4D dataset.}}
\vspace{-0.2cm}
\resizebox{0.75\columnwidth}{!}{
\begin{tabular}{l | c c c}
\toprule
Methods & Precision & EDA & AUC \\
\midrule
Random & 49.9 & 49.7 & 51.2 \\
EgoPED \cite{lee2024error} & 56.5 & 69.8 & 54.9 \\
\midrule
AMNAR (Ours) & \textbf{66.8} & \textbf{72.3} & \textbf{60.2} \\
\bottomrule
\end{tabular}
}
\label{tab:error_detection_captaincook4d}
\vspace{-10pt}
\end{table}

\subsection{Comparisons with SoTA Methods}
\vspace{-0.1cm}
\label{sec:results_and_comparison}
% In this section, we present the results of our experiments conducted on the EgoPER\cite{lee2024error}, HoloAssist\cite{wang2023holoassist} and CaptainCook4D\cite{peddi2023captaincook4d} datasets, comparing our proposed method, AMNAR, against state-of-the-art (SOTA) error detection approaches.

We compare our AMNAR with SoTA error detection approaches \cite{lee2024error,liu2021hybrid,ristea2022self,wu2022self} on the EgoPER\cite{lee2024error}, HoloAssist\cite{wang2023holoassist} and CaptainCook4D\cite{peddi2023captaincook4d} datasets.

As shown in \cref{tab:error_detection_egoper}, our results on the EgoPER dataset outperform all methods, with AMNAR achieving an average \textbf{7.4\%} improvement in EDA and \textbf{6.5\%} in AUC over EgoPED \cite{lee2024error}. For the HoloAssist dataset (\cref{tab:error_detection_results_in_holoassist}), AMNAR improves EDA by \textbf{3.7\%} and AUC by \textbf{1.3\%}. On the CaptainCook4D dataset (\cref{tab:error_detection_captaincook4d}), AMNAR surpasses EgoPED with improvements of \textbf{10.3\%} in Precision, \textbf{2.5\%} in EDA, and \textbf{5.3\%} in AUC, demonstrating robustness across diverse procedural tasks with complex action sequences.

\begin{table*}[t]
\centering
\caption{\textbf{Ablation studies on the Potential Action Prediction Block (PAPB) and the Representation Reconstruction Block (RRB).} ``w/o PAPB \& RRB'' excludes both PAPB and RRB, using only previously executed action features with a local self-attention mechanism. ``Random Selection'' indicates a variant where candidate actions are randomly selected instead of using PAPB to predict valid next actions.}
\vspace{-0.2cm}
\resizebox{\textwidth}{!}{ % Resize table to fit page width
\begin{tabular}{l|cc |cc cc cc cc cc| cc}
\toprule
\multirow{2}{*}{Variants} & \multicolumn{2}{c|}{Components} & \multicolumn{2}{c}{Quesadilla} & \multicolumn{2}{c}{Oatmeal} & \multicolumn{2}{c}{Pinwheel} & \multicolumn{2}{c}{Coffee} & \multicolumn{2}{c|}{Tea} & \multicolumn{2}{c}{All} \\
 & PAPB & RRB & EDA & AUC & EDA & AUC & EDA & AUC & EDA & AUC & EDA & AUC & EDA & AUC \\
\midrule
w/o PAPB \& RRB & \xmark & \xmark & 58.7 & 68.7 & 58.9 & 68.4 & 55.5 & 56.6 & 67.6 & 64.7 & \textbf{59.5} & 61.5 & 60.0 & 64.0 \\
Random Selection & \xmark & \cmark & 56.9 & 67.2 & 60.1 & 72.5 & 50.2 & 56.2 & 71.7 & 61.7 & 54.1 & 57.3 & 58.6 & 63.0 \\
AMNAR & \cmark & \cmark & \textbf{61.4} & \textbf{71.9} & \textbf{65.0} & \textbf{75.4} & \textbf{65.0} & \textbf{65.4} & \textbf{73.5} & \textbf{67.8} & 57.0 & \textbf{61.9} & \textbf{64.4} & \textbf{68.5} \\
\bottomrule
\end{tabular}
} % End resizebox
\vspace{-0.3cm}
\label{tab:aipm_comparison}
\end{table*}

\begin{table*}[t]
\centering
\caption{\textbf{Ablation studies across six configurations}: action representation types, prediction methods, video features, distance metrics, training sample selection strategies, task graph construction. The \colorbox{green!15}{highlighted} rows indicate our default implementation.}
\vspace{-0.2cm}
\begin{subtable}{0.25\textwidth}
    \centering
    \small
    \caption{Action Representation Types}
    \begin{tabular}{lcc}
        \toprule
        Variants & EDA & AUC \\
        \midrule
        Text-Based & \textbf{65.6} & 64.1 \\
        \rowcolor{green!15} Cluster-Centered & 64.4 & \textbf{68.5} \\
        \bottomrule
    \end{tabular}
\end{subtable}
\hspace{0.03\textwidth}
\begin{subtable}{0.31\textwidth}
    \centering
    \small
    \caption{Prediction Methods}
    \begin{tabular}{lcc}
        \toprule
        Variants & EDA & AUC \\
        \midrule
        Direct Prediction & 64.2 & 62.2 \\
        \rowcolor{green!15} Cluster Center Residual & \textbf{64.4} & \textbf{68.5} \\
        \bottomrule
    \end{tabular}
\end{subtable}
\hspace{0.03\textwidth}
\begin{subtable}{0.3\textwidth}
    \centering
    \small
    \caption{Visual Features}
    \begin{tabular}{lcc}
        \toprule
        Variants & EDA & AUC \\
        \midrule
        DINOv2 Feature & \textbf{71.0} & \textbf{69.2} \\
        \rowcolor{green!15} I3D Feature & 64.4 & 68.5 \\
        \bottomrule
    \end{tabular}
\end{subtable}

\vspace{0.1cm}

\begin{subtable}{0.25\textwidth}
    \centering
    \small
    \caption{Distance Metrics}
    \begin{tabular}{lcc}
        \toprule
        Variants & EDA & AUC \\
        \midrule
        L1 Norm & 62.4 & 63.9 \\
        Cosine Similarity & 63.2 & 62.9 \\
        \rowcolor{green!15} Euclidean & \textbf{64.4} & \textbf{68.5} \\
        \bottomrule
    \end{tabular}
\end{subtable}
\hspace{0.03\textwidth}
\begin{subtable}{0.31\textwidth}
    \centering
    \small
    \caption{Training Sample Selection Strategies}
    \begin{tabular}{lcc}
        \toprule
        Variants & EDA & AUC \\
        \midrule
        Training on GT Segments & \textbf{69.3} & 51.4 \\
        Training on ASM Output & 65.1 & 65.0 \\
        \rowcolor{green!15} Hybrid Training & 64.4 & \textbf{68.5} \\
        \bottomrule
    \end{tabular}
\end{subtable}
\hspace{0.03\textwidth}
\begin{subtable}{0.3\textwidth}
    \centering
    \small
    % \vspace{0.2cm}
    \caption{Task Graph Construction}
    \begin{tabular}{lcc}
        \toprule
        Variants & EDA & AUC \\
        \midrule
        Training Set Graph & \textbf{66.6} & 66.1 \\
        \rowcolor{green!15} Predefined Graph & 64.4 & \textbf{68.5} \\
        \bottomrule
    \end{tabular}
\end{subtable}
\label{tab:ablation_study_compact}
\vspace{-0.5cm}
\end{table*}

\subsection{Ablation Studies}
\vspace{-0.1cm}
\label{sec:effectiveness_of_AMNAR_component}
In this section, we conduct a series of \textbf{ablation studies} on the EgoPER dataset \cite{lee2024error} to evaluate the contributions of each component in our AMNAR framework. 
% By isolating and evaluating these components, we provide a detailed analysis of how each component influences error detection performance.

\noindent\textbf{Potential Action Prediction Block (PAPB).}  
The PAPB identifies valid next actions in procedural tasks, enabling AMNAR to model multiple normal action representations post-sequence, as outlined in the method section. We evaluate its role by comparing ``AMNAR'' with ``Random Selection'', a variant using random action selection instead of PAPB’s contextual prediction. As shown in \cref{tab:aipm_comparison}, AMNAR with PAPB boosts EDA by 5.8\% and AUC by 5.5\% over the random variant, proving that task-informed prediction enhances error detection in complex workflows.

\vspace{0.1cm}
\noindent\textbf{Representation Reconstruction Block (RRB).}  
The RRB generates context-aware normal action representations for PAPB-predicted actions, as described earlier. We test its impact by comparing ``AMNAR'' against ``w/o PAPB \& RRB'', which omits RRB and uses past action features with self-attention. \cref{tab:aipm_comparison} shows AMNAR with RRB improves EDA by 4.4\% and AUC by 4.5\% over this variant, confirming RRB’s role in enhancing error detection through tailored representations.

\vspace{0.1cm}
\noindent\textbf{Action Representation Types.}
AMNAR uses cluster-centered action representations to model the diversity of valid next actions by dynamically capturing their average distribution within each action class. Compared to text-based embeddings (e.g., BERT \cite{devlin2018bert}), our ablation study (\cref{tab:ablation_study_compact} (a)) shows superior performance, likely due to better alignment with the model’s feature space, enhancing robustness to subtle execution errors.

\vspace{0.1cm}
\noindent\textbf{Prediction Methods.} 
We compare direct prediction, which generates full features from context, with residual-based prediction, where the model predicts deviations from a cluster center for each intention-aligned action class. Per \cref{tab:ablation_study_compact} (b), the residual method boosts EDA by 0.2\% and AUC by 6.3\%. It enhances the learning of action-context relationships and sensitivity to subtle deviations from normal action distributions.

\vspace{0.1cm}
\noindent\textbf{Distance Metrics.}
We compare \(L_1\)-norm, cosine similarity, and Euclidean distance for feature alignment. As shown in \cref{tab:ablation_study_compact} (d), Euclidean distance outperforms others, achieving an EDA of 64.4\% and AUC of 68.5\%, against cosine similarity (63.2\% EDA, 62.9\% AUC) and \(L_1\)-norm (62.4\% EDA, 63.9\% AUC). Its sensitivity to magnitude and direction offers the best balance for error detection in complex procedural tasks.

\begin{figure*}[t]
    \centering
    % \vspace{-0.3cm}
    \includegraphics[width=0.75\linewidth]{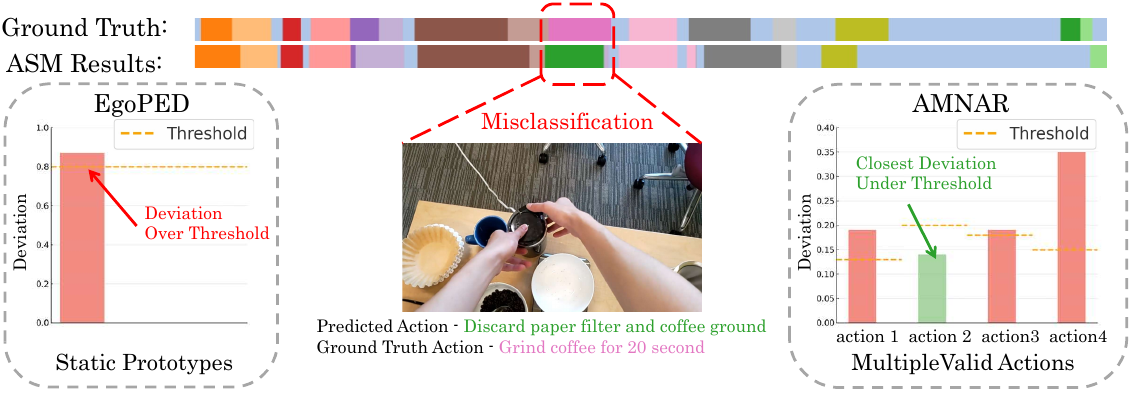} 
    \vspace{-0.3cm}
    \caption{Error detection when the Action Segmentation Model (ASM) misclassifies an action. AMNAR correctly identifies the action as normal. In contrast, the EgoPED framework incorrectly detects a false positive.}
    \label{fig:visualization}
    \vspace{-0.6cm}
\end{figure*}

\vspace{0.1cm}
\noindent\textbf{Training Sample Selection Strategies.} 
Effective training samples are key to robust error detection. We assess three approaches: (1) GT Segments, using ground-truth data for high accuracy but less resilience to ASM noise; (2) ASM Output, leveraging unfiltered ASM predictions for deployment alignment, risking label errors; and (3) Hybrid Training (\cref{sec:training_strategy_and_objective_function}), filtering ASM segments by an overlap threshold (\(\tau\)) with GT, balancing accuracy and ASM consistency. Per \cref{tab:ablation_study_compact} (e), Hybrid achieves the best AUC (68.5\%) with a solid EDA (64.4\%), optimizing robustness and precision.

\vspace{0.1cm}
\noindent\textbf{Visual Features.} 
Our AMNAR utilizes clip-level features from I3D for fair comparisons with existing works\cite{lee2024error}. Besides, Fig.S3 (2nd row, rightest) of Supplementary shows that AMNAR can detect fine-grained errors with I3D features. Moreover, using more robust feature representations further helps error detection. To highlight this, we replace the I3D with DINOv2, and the results are in \cref{tab:ablation_study_compact} (c).

\vspace{0.1cm}
\noindent\textbf{Task Graph Construction.}
Following the previous works\cite{ding2023every,lee2024error,seminara2024differentiable}, AMNAR uses predefined task graphs.\textbf{Note that we can also construct task graphs from the training set.}
\cref{tab:ablation_study_compact} (f) shows that the impact of constructing the graph from the training set is mild. 
Also, experiments on HoloAssist (\cref{tab:error_detection_results_in_holoassist}) and CaptainCook4D (\cref{tab:error_detection_captaincook4d}) utilize task graphs constructed from the training set and gain significant performance improvement.

% \vspace{-0.4cm}
% \subsection{Influence of Previous Errors}
% To assess whether previous errors affect subsequent normal action reconstruction---given our reliance on temporal context---we performed an ablation study. Per \cref{tab:influence_previous_errors}, multiple errors mildly reduce error detection performance. The table contrasts the ``w/o previous errors'' group (first error and prior actions) with the ``w previous errors'' group (subsequent actions). The former yields higher EDA due to fewer errors, yet both show similar results.

\subsection{Influence of Previous Errors}
\vspace{-0.1cm}
In procedural tasks, errors can accumulate over time. Since our method relies on temporal context from previous actions, a natural question could be raised: \emph{How do errors in previous actions affect the future reconstruction of subsequent normal action representations?} 

To address this, we conduct an ablation study to evaluate the robustness of AMNAR when prior errors are present in the action sequence. As shown in \cref{tab:influence_previous_errors}, we divide the test samples into two groups: the ``w/o previous errors'' group, which includes the first error in a sequence along with prior normal actions, and the ``w previous errors'' group, which covers actions following one or more prior errors. The results indicate that multiple prior errors cause a mild reduction in error detection performance. The ``w/o previous errors'' group achieves a higher EDA (70.8\%) compared to the ``w previous errors'' group (64.3\%), likely due to fewer distortions in the temporal context. However, the AUC remains comparable (67.3\% vs. 68.0\%), suggesting that AMNAR retains reasonable robustness even when prior errors are present. This demonstrates that, while prior errors introduce some challenges, our adaptive representation strategy mitigates their impact effectively.

\begin{table}[t]
\centering
\caption{\textbf{Influence of previous errors.} Previous errors cause only a mild reduction in error detection performance.}
\vspace{-0.3cm}
\resizebox{0.6\columnwidth}{!}{
\begin{tabular}{l|cc}
    \toprule
    Variants & EDA & AUC \\
    \midrule
    w/o previous errors & \textbf{70.8} & 67.3 \\
    w previous errors & 64.3 & \textbf{68.0} \\
    \bottomrule
\end{tabular}
}
\label{tab:influence_previous_errors}
\vspace{-0.3cm}
\end{table}

\begin{table}[t]
\centering
\caption{\textbf{EDA of non-deterministic actions in EgoPER dataset.} AMNAR consistently outperforms baselines, excelling in handling complex, non-deterministic action sequences.}
\vspace{-0.2cm}
\resizebox{\columnwidth}{!}{
\begin{tabular}{l |c c c c c c}
\toprule
Methods\&Variants & Quesadilla & Oatmeal & Pinwheel & Coffee & Tea & All \\
\midrule
EgoPED\cite{lee2024error} & \textbf{74.0} & 65.5 & 56.5 & 65.3 & 64.8 & 65.2 \\
AMNAR (w/o PAPB \& RRB) & 60.2 & 61.4 & 61.3 & 72.5 & 63.4 & 63.8 \\
AMNAR (Random) & 57.9 & 63.8 & 61.7 & 71.5 & 57.6 & 62.5 \\
AMNAR & 73.8 & \textbf{75.5} & \textbf{66.8} & \textbf{76.7} & \textbf{75.6} & \textbf{73.7} \\
\bottomrule
\end{tabular}
\vspace{-1.5cm}
}
\label{tab:acc_non_deterministic_actions}
\end{table}

\subsection{Ability of Handling Multiple Valid Next Actions}
\vspace{-0.1cm}
% Our evaluation highlights the capability of AMNAR in handling multiple valid next actions, with the ``coffee'' task showing the highest frequency of such scenarios (see Supplementary B.2 \& B.3). Per \cref{tab:error_detection_egoper}, AMNAR boosts EDA by 18.2\% and AUC by 9.5\% for this task, exceeding average gains of 7.4\% and 6.5\%. Testing on non-deterministic actions—steps with multiple valid priors (see Supplementary B.2 \& B.3)—shows AMNAR achieving a top EDA of 73.7\% overall and 76.7\% for ``coffee'' (\cref{tab:acc_non_deterministic_actions}), outperforming EgoPED and variants, highlighting its prowess in complex, non-deterministic sequences.
% Our AMNAR framework addresses the challenge of error detection with multiple valid next actions. 
We provide more discussions about our AMNAR on the ability of handling multiple valid next actions.  
On the one hand, on the task of ``coffee'', which contains the most complex action branching patterns, our AMNAR achieves significant performance improvement of EDA by 18.2\% and AUC by 9.5\% according to \cref{tab:error_detection_egoper}. 
% To further validate its performance in handling such challenge, we selected the ``coffee'' task, which contains the most complex action branching patterns (see Supplementary B.2 \& B.3 \ym{for what}). 
% Analysis of this task reveals that AMNAR boosts EDA by 18.2\% and AUC by 9.5\% according to \cref{tab:error_detection_egoper}, surpassing the average improvements of 7.4\% and 6.5\%. 

On the other hand, we conduct an evaluation on those non-deterministic actions, which are preceded by actions with multiple valid next options.
Results in \cref{tab:acc_non_deterministic_actions} demonstrate AMNAR achieves a top EDA of 73.7\% overall and 76.7\% for ``coffee'', outperforming EgoPED and its variants, underscoring its strength in managing complex, non-deterministic sequences. Details of frequency analysis of multiple valid next actions and definition of non-deterministic actions are in Supplementary B.4 \& B.5.

\subsection{Discussions on Action Segmentation Module}
\vspace{-0.1cm}
% AMNAR and EgoPED share the same ASM for fair comparison. Though our focus is Error Detection (ED), not Action Segmentation (AS), we report AS metrics in \cref{tab:action_segmentaion}, where AMNAR outperforms EgoPED due to distinct feature use—EgoPED alters features via clustering and contrastive learning, while AMNAR leverages them as reconstruction input. Notably, \cref{tab:action_segmentaion} shows EgoPED with AMNAR’s segmentation results (denoted as EgoPED*) still lags in ED, proving AMNAR’s ED superiority stems from its novel design, not just better AS.
We further present the Action Segmentation (AS) results. Although AMNAR and EgoPED employ the same AS module for fair comparison, AMNAR consistently outperforms EgoPED, as shown in \cref{tab:action_segmentaion}. This performance gap arises due to differences in feature utilization—EgoPED modifies features through clustering and contrastive learning, whereas AMNAR uses them as input for reconstruction. Notably, even when EgoPED leverages AMNAR’s segmentation results (denoted as EgoPED* in \cref{tab:action_segmentaion}), it still underperforms in Error Detection (ED). This demonstrates that AMNAR’s superior ED performance stems from its innovative design, beyond mere improvements in AS.

\subsection{Visualization}
\vspace{-0.1cm}
As shown in \cref{fig:visualization}, a ``coffee'' task sample demonstrates resilience of AMNAR to ASM misclassification. The ground truth action, “Grind coffee for 20 seconds,” is erroneously labeled by ASM as “Discard paper filter and coffee grounds.” EgoPED, hindered by an incorrect prototype, misclassifies it as an error. In contrast, AMNAR employs multiple normal action representations, selecting the top matching one to accurately classify the action as normal. This underscores robustness of AMNAR in handling label ambiguity and varied action sequences. More visualization samples are in Section C of supplementary material.

\begin{table}[t]
    % \caption{\textbf{Results of action segmentation (Left) and Error Detection result of using better action segmentations in EgoPED (Right).} EgoPED* denotes EgoPED with AMNAR’s segmentation results.}
    % \caption{\textbf{Action segmentation performance comparison (Left). Error detection results for EgoPED* (EgoPED using AMNAR’s segmentation) and AMNAR (Right).}}
    \caption{\textbf{Analysis of Action Segmentation Module.} \textbf{Left}: Our AMNAR outperforms EgoPED on action segmentation task. \textbf{Right}: EgoPED under-performs our AMNAR even if used with action segmentation results of AMNAR (noted it by *).}
    \vspace{-10pt}
    \centering
    \scriptsize
    \begin{subtable}{0.5\linewidth}
      \centering
      % \caption{Action Segmentation Results}
      \begin{tabular}{@{\hspace{4pt}}l|@{\hspace{4pt}}c@{\hspace{4pt}}c@{\hspace{4pt}}c@{\hspace{4pt}}c@{\hspace{4pt}}}
        \toprule
        Methods & IoU & Edit & F1@0.5 & Acc \\
        \midrule
        EgoPED & 44.6 & 61.3 & 47.5 & 68.5 \\
        \textbf{AMNAR} & \textbf{56.3} & \textbf{69.4} & \textbf{57.3} & \textbf{75.3} \\
        \bottomrule
      \end{tabular}
    \end{subtable}%
    \begin{subtable}{0.5\linewidth}
      \centering
      % \caption{EgoPED w better Segmentation}
      \begin{tabular}{@{\hspace{4pt}}l|@{\hspace{4pt}}c@{\hspace{4pt}}c@{\hspace{4pt}}}
        \toprule
        Methods & Avg. EDA & Avg. AUC \\
        \midrule
        EgoPED* & 63.1 & 61.9 \\
        \textbf{AMNAR} & \textbf{64.4} & \textbf{68.5} \\
        \bottomrule
      \end{tabular}
    \end{subtable}
    \vspace{-0.6cm}
    \label{tab:action_segmentaion}
\end{table}

\vspace{-0.1cm}
\section{Conclusion}
\vspace{-0.1cm}
\label{sec:conclusion}
% We introduce the Adaptive Multiple Normal Action Representation (AMNAR) framework, which excels in error detection for procedural tasks by dynamically modeling multiple valid next actions and their normal representations. Outperforming state-of-the-art methods on the EgoPER, HoloAssist and CaptainCook4D datasets—particularly in tasks with diverse next steps—AMNAR proves that adaptive representation of multiple normal actions enhances accuracy in complex scenarios.

In this work, we uncover a critical limitation in existing error detection approaches: their inability to effectively handle scenarios of multiple valid next actions. To address this, we develop the Adaptive Multiple Normal Action Representation (AMNAR) framework, which dynamically predicts and reconstructs representations for all valid next actions. Through comprehensive experiments—including comparative analyses, ablation studies, and evaluations of non-deterministic actions across three datasets—we confirm the effectiveness of AMNAR. We believe this adaptive, multi-representation strategy could improve error detection and contribute to advancements in broader action understanding fields.

\section*{Acknowledgement}
This work was supported partially by NSFC(92470202, U21A20471), National Key Research and Development Program of China (2023YFA1008503), Guangdong NSF Project (No. 2023B1515040025). The authors thank anonymous reviewers and ACs for their constructive suggestions.

{
    \small
    \bibliographystyle{ieeenat_fullname}
    \bibliography{main}
}

% WARNING: do not forget to delete the supplementary pages from your submission 
\clearpage
\appendix

\section*{Overview}

In this supplementary material, we provide the following sections:

\begin{itemize}
    \item \cref{method_details}: More details about our proposed Adaptive Multiple Normal Action Representation (AMNAR) framework.
    \item \cref{experimental_setup}: More details about the experiment setups.
    \item \cref{visualization}: More visualizations for further demonstration on the effectiveness of the proposed AMNAR.
\end{itemize}

\section{More Details about AMNAR}
\label{method_details}

\subsection{Potential Action Prediction Block}

% The Potential Action Prediction Block (PAPB) is a key component designed to predict all potential next actions based on the task graph \( G \) and the executed action sequence \( s \). The variable-definition reference table for PAPB is shown in \cref{tab:variables}, and the pseudocode for PAPB is presented in \cref{alg:papb_short}.

The Potential Action Prediction Block (PAPB) is a key component designed to predict all potential next actions based on the task graph \( G \) and the executed action sequence \( s \). The variable-definition reference table and pseudocode for PAPB are shown in \cref{tab:variables} and \cref{alg:papb_short}, respectively.

\vspace{0.1cm}
\noindent{\textbf{Adjacency List Construction.}}
PAPB begins by converting the task graph \( G \) into an adjacency list \( A \), where each node in the graph links to its direct successors.

\vspace{0.1cm}
\noindent{\textbf{Longest Subsequence Identification.}}
PAPB employs dynamic programming to find the longest subsequence \( s^* \) in \( s \) that adheres to the relationships defined by \( G \). The algorithm maintains two tables: \( \text{subseq}[i] \), which stores the longest non-branching subsequence ending at index \( i \), and \( \text{dp}[i] \), which stores the \( \text{subseq}[i] \).  A \textbf{non-branching subsequence} is defined as a sequence of nodes that form a continuous path in the task graph \( G \), where all nodes are connected sequentially without any splits or branches (e.g., [0, 1, 2] in \cref{fig:task_graph}).

For each action \( y_i \) in \( s \), the algorithm iterates over all previous actions \( y_j \) (where \( j < i \)) and checks whether \( y_i \) and \( y_j \) are connected in the task graph \( G \). If this condition is met, \( \text{dp}[i] \) and \( \text{subseq}[i] \) are updated as follows:

\begin{equation}
\text{dp}[i] = \max(\text{dp}[i], \text{dp}[j] + 1),
\end{equation}
\begin{equation}
\resizebox{0.9\linewidth}{!}{$
\text{subseq}[i] =
\begin{cases} 
    \text{subseq}[j] \cup \{y_i\}, & \text{if } \text{dp}[j] + 1 > \text{dp}[i], \\
    \text{subseq}[i] \cup (\text{subseq}[j] \cup \{y_i\}), & \text{if } \text{dp}[j] + 1 = \text{dp}[i].
\end{cases}
$}
\end{equation}

After processing \( s \), the algorithm identifies the maximum value in \( \text{dp} \), locating the index \( k \) with the longest non-branching subsequence $L$.

\vspace{0.1cm}
\noindent{\textbf{Merging Connected Nodes.}}
While the longest subsequence identified in dynamic programming represents a non-branching path (e.g., [0, 1, 2] in \cref{fig:task_graph}), it may not capture all executed actions in scenarios where multiple branches exist in the task graph. To address this, PAPB iteratively examines each subsequence. For each subsequence, if any of its nodes matches a node in \( L \), the subsequence is considered connected to \( L \), and its nodes are merged into \( L \). This merging process ensures that \( L \) includes all nodes relevant to the executed actions, resulting in the complete merged sequence \( s^* \), which accurately reflects all executed actions within the task graph.

\begin{figure}[t]
    \centering
    \includegraphics[width=\linewidth]{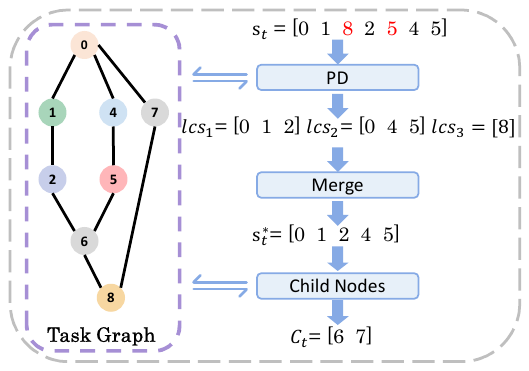} 
    \caption{The Potential Action Prediction Block (PAPB) derives the longest matching subsequence from the executed sequence using the task graph. This subsequence is then used to identify all reachable nodes, representing valid next actions. This figure is reproduced from the main text for reference.}
    \label{fig:task_graph}
    \vspace{-10pt}
\end{figure}

\vspace{0.1cm}
\noindent \textbf{Next Action Prioritization.}
Based on \( s^* \), PAPB computes the set of potential next actions \( PA \) as:
\begin{equation}
PA = (\bigcup_{a \in \text{$s^*$}} A[a]) \setminus \text{$s^*$}.
\end{equation}
In this formula, \( A[a] \) represents the set of direct successors of node \( a \) in the task graph \( G \), as derived from the adjacency list. By iterating over all nodes \( a \) in the longest merged subsequence \( s^* \), the union \( \bigcup_{a \in \text{$s^*$}} A[a] \) aggregates the successors of all nodes in \( s^* \). The subtraction \( \setminus \text{$s^*$} \) ensures that only actions not already included in \( s^* \) are retained in \( PA \). This guarantees that \( PA \) contains all valid next actions that can logically follow the executed actions, without duplication.

PAPB efficiently combines dynamic programming and graph traversal to provide actionable insights from \( s \) and \( G \). For detailed implementation, refer to \cref{alg:papb_short}.

\begin{algorithm}[H]
\caption{Potential Action Prediction Block (PAPB)}
\label{alg:papb_short}
\begin{algorithmic}
    \STATE \textbf{Input:} Task graph $G$, Executed action sequence $s$
    \STATE \textbf{Output:} Prioritized list of next actions $PA$
    
    \STATE \textcolor{grassgreen}{\textbf{\# Build Adjacency Lists:}}
    \STATE Initialize $A[u] = \emptyset$ for all $u \in G$
    \FOR{each edge $(u, v)$ in $G$}
        \STATE $A[u] \gets A[u] \cup \{v\}$
    \ENDFOR
    
    \STATE \textcolor{grassgreen}{\textbf{\# DP Process:}}
    \STATE Initialize $\text{dp}[i] \gets 1$ and $\text{subseq}[i] \gets \{y_i\}$ for all $i$
    
    \FOR{$i \gets 1$ to $n$}
        \FOR{$j \gets 1$ to $i-1$}
            \IF{$y_i \in A[y_j]$ \textbf{or} $y_j \in A[y_i]$}
                \IF{$\text{dp}[j] + 1 > \text{dp}[i]$}
                    \STATE $\text{dp}[i] \gets \text{dp}[j] + 1$
                    \STATE $\text{subseq}[i] \gets \text{subseq}[j] \cup \{y_i\}$
                \ELSIF{$\text{dp}[j] + 1 == \text{dp}[i]$}
                    \STATE $\text{subseq}[i] \gets \text{subseq}[i] \cup \text{subseq}[j] \cup \{y_i\}$
                \ENDIF
            \ENDIF
        \ENDFOR
    \ENDFOR
    
    \STATE \textcolor{grassgreen}{\textbf{\# Collect Max-Length Subsequences:}}
    \STATE $k \gets \max(\text{dp}[1], \text{dp}[2], \dots, \text{dp}[n])$
    \STATE $L \gets \bigcup \{\text{subseq}[i] \mid \text{dp}[i] = k\}$
    
    \STATE \textcolor{grassgreen}{\textbf{\# Merge Connected Nodes in $L$:}}
    \STATE Initialize $s^* \gets L$
    \FOR{each $node$ in $L$}
        \FOR{each $neighbor \in A[node]$}
            \IF{$neighbor \in L$}
                \STATE $s^* \gets s^* \cup \{neighbor\}$
            \ENDIF
        \ENDFOR
    \ENDFOR
    
    \STATE \textcolor{grassgreen}{\textbf{\# Collect Potential Next Actions:}}
    \STATE $PA \gets (\bigcup_{a \in s^*} A[a]) \setminus s^*$
    
    \STATE \textbf{Return} $PA$
\end{algorithmic}
\end{algorithm}

\begin{table}[h]
\caption{Variable Definitions of PAPB}
\label{tab:variables}
\centering
\resizebox{\linewidth}{!}{%
\begin{tabular}{ll}
\toprule
\textbf{Variable} & \textbf{Definition} \\
\midrule
$G$ & Task graph\\
$s$ & Executed action sequence \\
$s^*$ & The longest matching subsequence \\
$A$ & Adjacency list of $G$ \\
$A[a]$ & The set of direct successors of node \( a \) \\
$\text{subseq}[i]$ & Longest non-branching subsequence ending at index $i$ \\
$\text{dp}[i]$ & Length of $\text{subseq}[i]$ \\
$k$ & Index with the maximum $\text{dp}[k]$ \\
$L$ & Longest non-branching subsequence \\
PA & Final potential next actions \\
\bottomrule
\end{tabular}%
}
\end{table}

\subsection{Representation Reconstruction Block}
\begin{figure}[t]
    \centering
    \includegraphics[width=0.65\linewidth]{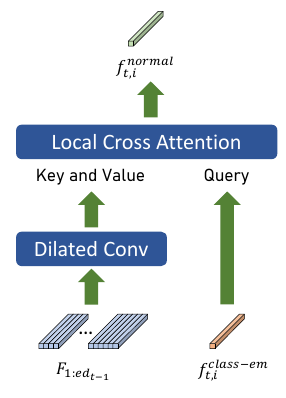} 
    \caption{Architecture of the Representation Reconstruction Block (RRB). The RRB reconstructs the $i$-th normal action representation \( f_{t,i}^{\text{normal}} \) for time \( t \) by combining the frame-wise refined features \( F_{1:ed_{t-1}} \) (key and value) and the action class embedding \( f_{t,i}^{\text{class-emb}} \) (query).}
    \label{fig:rrb}
    \vspace{-10pt}
\end{figure}

The Representation Reconstruction Block (RRB) is designed to reconstruct multiple normal action representations at time \( t \) using the frame-wise features of executed actions and the embedding of the \( t \)-th action. The RRB consists of two key components: a dilated convolutional layer and a local cross-attention module, as illustrated in \cref{fig:rrb}. 

To ensure temporal causality, all modules within the RRB are implemented in a causal manner. Specifically, when reconstructing the normal action representations at time \( t \), the frame-wise features corresponding to time \( t \) and any future frames are not accessible, thereby adhering to the sequential nature of the task.

\vspace{0.1cm}
\noindent \textbf{Dilated Convolutional Layer.} 
The dilated convolutional layer employs a kernel size of 3 and consists of 5 layers. The dilation rate of the first layer is set to 1, while the subsequent layers follow an exponential growth pattern. Specifically, the dilation rate \( d_i \) for the \( i \)-th layer is defined as:
\begin{equation}
d_i = 3^i.
\end{equation}
This design allows the receptive field to expand exponentially with depth.

\vspace{0.1cm}
\noindent \textbf{Local Cross Attention.} 
The local cross attention module  consists of a single attention layer with a local window length of 32 and 2 attention heads. Depthwise convolutions project the query, key, and value features, with causal padding ensuring only past and current time steps are accessible, preserving temporal causality.

\vspace{0.1cm}
\noindent \textbf{Action Class Embedding.} 
As mentioned in Section \textcolor{red}{3.3} of the main text, \( F_{1:ed_{t-1}} \) represents the frame-wise refined visual features extracted from the Action Segmentation Model up to frame \( ed_{t-1} \). The \( f^{\text{class-emb}}(y) \) represents the class embedding for action class \( y \), computed as the mean feature of all action samples belonging to this class. Formally, it is defined as:

\begin{equation}
f^{\text{class-emb}}(y) = \frac{\sum_{t \in \mathcal{I}_y} f^{\text{action}}_t}{N_y},
\end{equation}
where \( \mathcal{I}_y \) is the set of indices for samples belonging to class \( y \), \( N_y = |\mathcal{I}_y| \) is the total number of samples in this class, and \( f^{\text{action}}_t \) represents the feature of the \( t \)-th action sample. This class embedding serves as a representative feature for action class \( y \).

The \( f^{\text{class-emb}}_{t,i} \) represents the class embedding for the $i$-th potential action class corresponding to the \( t \)-th action. It is used as the query input in the Local Cross Attention module (see \cref{fig:rrb}), where it interacts with the key and value features derived from the frame-wise refined features \( F_{1:ed_{t-1}} \) after processing through the dilated convolution layer.

\section{More Experimental Setups}
\label{experimental_setup}
In this section, we provide comprehensive details about the experimental setup to complement the descriptions in the main text. Specifically, we elaborate on the preprocessing and usage of the HoloAssist\cite{wang2023holoassist} datasets, frequency analysis of multiple valid next actions, as well as the experimental environment and hyperparameter settings.

\subsection{HoloAssist Dataset}
Since the official release of the HoloAssist dataset lacks a designated test set, we train our AMNAR and EgoPED \cite{lee2024error} frameworks on the training set, compute thresholds using the training set, and evaluate performance on the validation set. The tasks used for training and validation, along with their respective durations, are summarized in Table~\ref{tab:holoassist_duration}. To train the Action Segmentation Model (ASM), we utilize the fine-grained action annotations, specifically either verb or noun labels, as segment labels.

The HoloAssist training set includes both normal and erroneous actions. To ensure accurate learning of normal action representations, we train AMNAR exclusively on normal actions, excluding erroneous ones during training. For HoloAssist experiments, due to the absence of an official test set, we follow a standard split by training on the provided training set (approximately 166 hours of video from 350 instructor-performer pairs) and evaluating on the validation set. Additionally, we exclude the “Belt” task from final evaluations, as it contains only one error-free sample, which could skew performance metrics.

Moreover, some action classes appear only in the validation set and are absent from the training set. To maintain consistency during inference, we classify these unseen classes as background actions. For task graph construction, since HoloAssist lacks predefined task graphs, we generate them by analyzing all training sequences.

We also introduce a random baseline for HoloAssist experiments. This baseline employs the same ASM trained with the aforementioned strategy and, during inference, randomly classifies each action segment as either normal or erroneous.

\begin{table}[t]
    \centering
    \caption{Duration of Training and Validation Sets for HoloAssist Tasks (in minutes)}
    \begin{tabular}{lcc}
    \toprule
    \textbf{Task Name}             & \textbf{Train (min)} & \textbf{Val (min)} \\
    \midrule
    atv                            & 84.63           & 12.37          \\
    circuitbreaker                 & 45.30           & 8.62           \\
    coffee                         & 137.17          & 16.38          \\
    computer                       & 226.43          & 38.95          \\
    dslr                           & 289.22          & 38.15          \\
    gladom\_assemble               & 320.95          & 50.60          \\
    gladom\_disassemble            & 211.03          & 29.02          \\
    gopro                          & 561.58          & 78.18          \\
    knarrevik\_assemble            & 843.08          & 114.08         \\
    knarrevik\_disassemble         & 465.00          & 71.63          \\
    marius\_assemble               & 357.58          & 52.28          \\
    marius\_disassemble            & 208.38          & 36.83          \\
    navvis                         & 122.65          & 21.25          \\
    nespresso                      & 225.47          & 28.47          \\
    printer\_big                   & 162.15          & 26.87          \\
    printer\_small                 & 295.05          & 42.32          \\
    rashult\_assemble              & 942.42          & 128.90         \\
    rashult\_disassemble           & 545.65          & 68.47          \\
    switch                         & 469.07          & 70.82          \\
    \bottomrule
    \end{tabular}
    \label{tab:holoassist_duration}
    \end{table}

\subsection{CaptainCook4D Dataset}
The CaptainCook4D dataset \cite{peddi2023captaincook4d} is a large-scale egocentric 4D dataset designed for understanding errors in procedural cooking activities. It comprises 384 recordings (94.5 hours) of individuals performing 24 different recipes in real kitchen environments. The dataset includes videos of participants correctly following recipe instructions as well as instances where they deviate and introduce errors. It provides 5.3K step annotations and 10K fine-grained action annotations, with errors categorized into seven distinct types. Data modalities include RGB video, depth, 3D hand joint tracking, and IMU data, captured using a head-mounted GoPro and HoloLens2.

For our experiments, since CaptainCook4D lacks predefined task graphs, we generate them by analyzing all training sequences, similar to the approach used for HoloAssist. To focus on execution-related errors, we exclude the “Missing Step” and “Ordering” error types during evaluation, as these sequence-level anomalies are beyond the primary scope of AMNAR.

\subsection{Task Graph Generation for Procedural Task Modeling}
To better model procedural tasks in both HoloAssist and CaptainCook4D, we derive task graphs from action sequences, as these datasets do not provide predefined graphs. Each task graph is represented as a Directed Acyclic Graph (DAG) that captures valid action transitions based on observed sequences.

The graph construction consists of three steps:
1. \textbf{Extract Action Sequences}: Identify non-background action sequences from the recordings and insert a start state (e.g., background) at the beginning of each sequence.
2. \textbf{Compute Transition Weights}: Measure the co-occurrence frequency of each action pair across all sequences to form a weighted transition matrix.
3. \textbf{Build a Maximum-Weight DAG}: Use a greedy algorithm to select the highest-weight edges while disallowing cycles, preserving only acyclic paths.

This procedure ensures that frequent, logically coherent transitions are included in the final task graph, providing a reliable structure for analyzing procedural tasks. 
For the complete pseudocode of this task graph generation process, please refer to \cref{alg:task_graph_gen}.

\begin{algorithm}[t]
    \caption{Task Graph Generation}
    \label{alg:task_graph_gen}
    \begin{algorithmic}
        \STATE \textbf{Input:} Action sequences $S$
        \STATE \textbf{Output:} Task graph $G$ as a list of edges
        
        \STATE \textcolor{grassgreen}{\textbf{\# Compute Transition Weights:}}
        \STATE Initialize $T[(u, v)] \gets 0$ for all possible $(u, v)$
        \FOR{each $seq \in S$}
            \FOR{$i \gets 0$ to $\text{len}(seq)-2$}
                \FOR{$j \gets i+1$ to $\text{len}(seq)-1$}
                    \STATE $T[(\text{seq}[i], \text{seq}[j])] \gets T[(\text{seq}[i], \text{seq}[j])] + 1$
                \ENDFOR
            \ENDFOR
        \ENDFOR
        
        \STATE \textcolor{grassgreen}{\textbf{\# Sort Transitions by Weight:}}
        \STATE $P \gets \text{sort}(T.\text{items}(), \text{key}=\text{weight}, \text{descending})$
        
        \STATE \textcolor{grassgreen}{\textbf{\# Build Maximum-Weight DAG:}}
        \STATE Initialize $G \gets \emptyset$
        \FOR{$(u, v)$ in $P$}
            \IF{adding $(u, v)$ to $G$ keeps $G$ acyclic}
                \STATE $G \gets G \cup \{(u, v)\}$
            \ENDIF
        \ENDFOR
        
        \STATE \textbf{Return} $G$
    \end{algorithmic}
\end{algorithm}

This approach ensures the task graph reflects frequent, logical action transitions while maintaining an acyclic structure, suitable for procedural task analysis.

\subsection{Frequency Analysis of Multiple Valid Next Actions}

In Section \textcolor{red}{4.4} of the main text, we compare average improvements across tasks, noting that the \textit{coffee} task has the highest occurrence of multiple valid next actions. This observation stems from a frequency analysis of multiple valid next actions using the following metrics: \textbf{non-deterministic action ratio}, \textbf{average number of valid next actions} and \textbf{average maximum transfer probability}.

A \textbf{non-deterministic action} is defined as an action whose preceding action has more than one potential next action. As illustrated in Figure~\ref{fig:task_graph}, consider action $a_1$, which follows action $a_0$. Since action $a_0$ has multiple potential next actions (actions $a_1$, $a_4$, $a_7$), action $a_1$ is considered a non-deterministic action (as are $a_4$ and $a_7$).

The \textbf{non-deterministic action ratio} refers to the proportion of non-deterministic actions among all actions within a task. A higher ratio indicates a greater prevalence of multiple valid next actions, contributing to task complexity. As shown in Table~\ref{tab:task_transition_metrics}, the tasks \textit{tea}, \textit{coffee}, and \textit{oatmeal} have notably high non-deterministic action ratios of 75.00\%, 70.59\%, and 69.23\%, respectively.

The \textbf{average number of valid next actions} represents the mean count of potential valid next actions for each action in a task. For instance, if action $a_0$ has potential next actions $a_1$, $a_2$, and $a_3$, the number of valid next actions is 3. A higher average indicates that actions generally have more possible subsequent actions, increasing the task's complexity. In terms of this metric, the \textit{coffee} task stands out with a value of 2.82, higher than those of other tasks.

The \textbf{average maximum transfer probability} is the average of the highest probabilities with which actions transition to their next actions. For example, if action $a_0$ transitions to $a_1$, $a_2$, and $a_3$ with probabilities of 20.00\%, 25.00\%, and 55.00\%, the maximum transfer probability for $a_0$ is 55.00\%. A lower average maximum transfer probability indicates greater uncertainty in transitioning to a specific next action, reflecting higher diversity in valid next steps. As shown in Table~\ref{tab:task_transition_metrics}, the \textit{coffee} and \textit{oatmeal} tasks have lower average maximum transfer probabilities of 67.27\% and 67.09\%, respectively.

The \textit{coffee} task stands out across all three metrics, indicating a high frequency of multiple valid next actions. This complexity makes it the most suitable task for demonstrating the effectiveness of our Adaptive Multiple Normal Action Representation (AMNAR) framework. Consistent with our frequency analysis, AMNAR achieves the most substantial improvement in error detection accuracy for the \textit{coffee} task, as evidenced in Table~1 of the main text. This correlation underscores the advantage of AMNAR in handling tasks with diverse and multiple valid action sequences.

\begin{table}[htbp]
\centering
\caption{Metrics for Task Transition Matrices Across Five Tasks. Higher non-deterministic ratios (\(\uparrow\)) indicate greater complexity due to multiple valid next actions. Higher average numbers of valid next actions (\(\uparrow\)) suggest increased complexity of a task. Lower average maximum transfer probabilities (\(\downarrow\)) indicate greater uncertainty in action transitions.}
\resizebox{\linewidth}{!}{%
\begin{tabular}{l|ccccc}
\toprule
\textbf{Metric} & \textbf{Tea} & \textbf{Coffee} & \textbf{Pinwheels} & \textbf{Oatmeal} & \textbf{Quesadilla} \\
\midrule
Non-deterministic Ratio (\%) \(\uparrow\)  & 75.00 & 70.59 & 26.67 & 69.23 & 40.00 \\
Avg. Valid Next Actions \(\uparrow\)        & 1.75  & 2.82  & 1.13  & 1.85  & 1.20  \\
Avg. Max Transfer Prob. (\%) \(\downarrow\) & 74.02 & 67.27 & 88.15 & 67.09 & 75.00 \\
\bottomrule
\end{tabular}%
}
\label{tab:task_transition_metrics}
\end{table}

\subsection{EDA of non-deterministic actions}
In Sec. \textcolor{red}{4.4} of the main paper, we evaluate the Error Detection Accuracy (EDA) of non-deterministic actions. This experiment measures the average frame-wise accuracy of non-deterministic actions in error detection.

\subsection{Experimental Environment and Hyperparameters}
All experiments are conducted on an Nvidia Tesla V100 GPU with 32GB of VRAM. The training process uses a batch size of 8 and runs for 200 epochs. The learning rate is initialized to 0.001 and adjusted dynamically using a cosine annealing schedule.

\section{Visualization Examples}
\label{visualization}

\cref{fig:visualization} presents additional visualization examples of error detection using the AMNAR framework on the EgoPER dataset \cite{lee2024error}. These examples highlight how AMNAR effectively identifies various types of errors in procedural tasks, demonstrating its robustness and adaptability in complex scenarios.

\begin{figure*}[t]
    \vspace{-195pt}
    % \centering
    \includegraphics[width=0.95\textwidth]{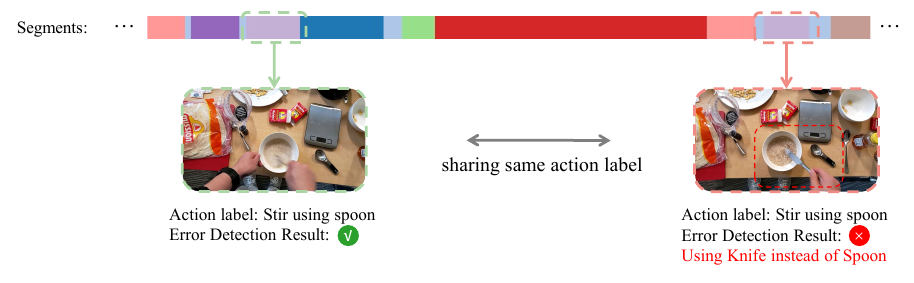} 
    \caption{In this example, the AMNAR framework encounters two actions sharing the same label: one is a correctly executed action (normal), and the other is an erroneous action. Despite the shared label, AMNAR successfully detects the error in the second action (using a knife instead of a spoon) while correctly identifying the first action as normal, avoiding any false positives.}
    \label{fig:normal_and_error_cropped}
\end{figure*}

As shown in \cref{fig:normal_and_error_cropped}, the AMNAR framework accurately detects errors even when they occur within actions sharing the same label, effectively distinguishing between normal and erroneous executions.

\begin{figure*}[t]
    \vspace{-20pt}
    \centering
    \includegraphics[width=0.95\textwidth]{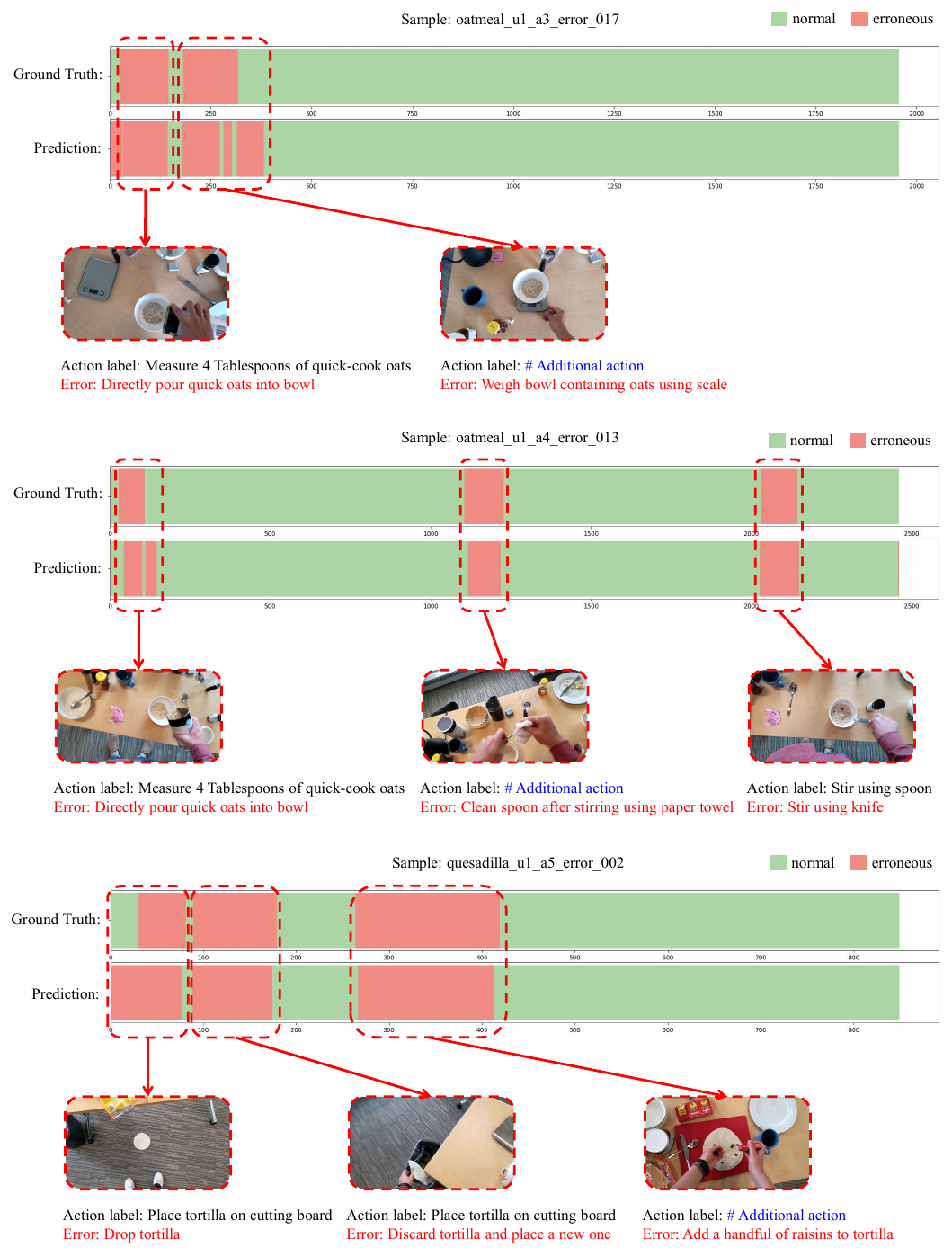} 
    \vspace{-8pt}
    \caption{Visualization examples from the EgoPER dataset using the AMNAR framework. In the \textbf{top sample}, two errors are detected: a misoperation—quick oats are poured directly into the bowl without measuring, and an additional action. The \textbf{middle sample} also contains three errors: a misoperation, an additional action, and using the wrong tool—stirring with a knife instead of a spoon. The \textbf{bottom sample} illustrates a sequence of errors: an accidental error—dropping the tortilla to the ground, followed by a corrective action—replacing the dropped tortilla with a new one, and finally an additional action—adding an incorrect ingredient.}
    \label{fig:visualization}
\end{figure*}

% \clearpage

\end{document}